\documentclass[11pt]{article}      

\usepackage{graphics} 

\usepackage{epsfig} 

\usepackage{bm} 
\usepackage{mathptmx} 

\usepackage{times} 
 
\usepackage{amsmath} 

\usepackage{amssymb}  

\usepackage{color}

\usepackage{algorithm}

\usepackage{algorithmic}

\newtheorem{dfn}{Definition}

\newtheorem{prop}{Proposition}

\newtheorem{rem}{Remark}

\newcommand{\RR}{{\mathbb R}}

\newcommand{\norm}[1]{\lVert#1\rVert}

\newcommand{\dotex}{{\frac{d}{dt}}}

\newcommand{\ba}{{\mathbf a}}

\newcommand{\bg}{{\mathbf g}}

\newcommand{\bb}{{\mathbf b}}

\newcommand{\boxi}{\bm{\xi}}
\newcommand{\bomega}{\bm{\omega}}
\newcommand{\bSigma}{\bm{\Sigma}}
\newcommand{\db}{\bm{\delta b}}
\newcommand{\bR}{{\bf R}}
\newcommand{\bd}{{\bf d}}

\newcommand{\bT}{{\mathbf T}}
\newcommand{\bV}{{\mathbf V}}
\newcommand{\bF}{{\mathbf F}}
\newcommand{\bX}{{\mathbf X}}
\newcommand{\bx}{{\mathbf x}}
\newcommand{\bv}{{\mathbf v}}

\newcommand{\bW}{{\mathbf W}}
\newcommand{\bU}{{\mathbf U}}

\newcommand{\bGamma}{\bm{\Gamma}}
\newcommand{\bUpsilon}{\bm{\Upsilon}}
\newcommand{\bbeta}{\bm{\eta}}

\title{\LARGE \bf A Mathematical Framework for IMU Error Propagation with Applications to Preintegration}

\author{Axel Barrau\footnote{ A. Barrau is with 
SAFRAN TECH, Groupe Safran, Rue des Jeunes Bois - Ch\^ateaufort, 78772 Magny Les Hameaux CEDEX and  MINES ParisTech, PSL Research University, Centre for robotics, 60 Bd St Michel 75006 Paris, France } \and Silv\`ere Bonnabel\footnote{S. Bonnabel is with MINES ParisTech, PSL Research University, Centre for robotics, 60 Bd St Michel 75006 Paris, France, and ISEA, University of New Caledonia, 98800  Noumea, New Caledonia
       {\tt\footnotesize [silvere.bonnabel]@mines-paristech.fr}}}

\date{}

\include{myLatex}

\graphicspath{{Figures/}}

\begin{document}

This is the online preprint of the paper ``A Mathematical Framework for IMU Error Propagation with Applications to Preintegration'',   published in the proceedings of ICRA 2020 - IEEE International Conference on Robotics and Automation.

\newpage

\maketitle
%
%
%
%


\begin{abstract}
To fuse information from inertial   measurement units (IMU) with other sensors one needs an accurate model for IMU error propagation in terms of position, velocity and orientation, a triplet we  call extended pose.  In this paper we leverage a  nontrivial result, namely  log-linearity  of inertial navigation equations based on the recently introduced Lie group $SE_2(3)$, to   transpose the recent methodology of Barfoot and Furgale for associating uncertainty with poses (position, orientation) of $SE(3)$ when using  noisy wheel speeds, to the case of  extended poses (position, velocity, orientation) of $SE_2(3)$ when using noisy IMUs. Besides, our approach to extended poses combined with log-linearity property  allows revisiting the theory of preintegration on manifolds and reaching a further theoretic level in this field.  We show  exact preintegration formulas that account for rotating Earth, that is, centrifugal force and Coriolis effect,  may  be derived as a byproduct. 
\end{abstract}

Geometric approaches to associating uncertainty with poses, essentially for mobile robot localization, have been quite   successful in the robotics community over the past decade. Since the discovery that mobile robots dispersion under the effect of sensor noise resembles more a ``banana" than a standard Gaussian ellipse, which can be traced back to \cite{thrun2000real},   studies   have evidenced the fact that the Lie group structure of the configuration space $SE(3)$ plays a prominent role in probabilistic robotics, see  \cite{barfoot2016state,chirikjian2014gaussian,barfoot2014associating,BayesianLieGroups2011,diemer2015invariant,barrau2013intrinsicp,hertzberg2013integrating,park2008kinematic}. In particular, Gaussian distributions in (Lie) exponential coordinates provide accurate approximations of banana distributions as first advocated by \cite{long2012banana}.  The reader is  also referred to the  recent monographs \cite{chirikjian2011stochastic,barfoot2016state}.

When  inertial sensors (gyrometers and accelerometers) embedded in an Inertial Measurement Unit  (IMU)  are utilized, one needs to manipulate extended poses, which are 9 dimensional elements that cannot be modeled as elements of $SE(3)$.  Moreover, the IMU propagation equations are not amenable to left multiplications on Lie groups of the form $\bT_{k+1}=\bT_k\bGamma_k$. This has  two consequences. First the results about uncertainty propagation of \cite{barfoot2014associating} are not easily transposed in an IMU context. Then, note  that if IMU propagation equations were of the form above we would readily  have $\bT_{k+N}=\bT_k(\Pi_k^{N-1}\bGamma_i)$, and preintegrating IMU measurements  as in \cite{forster2017manifold}  would be trivial, which is not the case. As a result,  the theory of preintegration on manifolds  \cite{forster2017manifold}  is more subtle and relies on smart algebraic tricks, see also \cite{martinelli2014closed,eckenhoff2019closed}. 

In \cite{barrau2015non,barrau2017invariant}  the   introduction of the group $SE_2(3)$,    along  with the discovery of the associated group affine property and hence log-linearity of IMU equations  using $SE_2(3)$,   proves a major step to overcome these obstacles. It has already led to more robust EKFs for data fusion with IMU, has prompted an industrial product  \cite{barrau2018invariant}, has improved EKF-based visual inertial consistency \cite{wu_invariant-ekf_2017, heo_consistent_2018,brossard2017unscented, caruso_2018, heo2018consistent} and robot state estimation \cite{Hartley-RSS-18,hartley2019contact}. 
 In this paper, we show the group  $SE_2(3)$  allows transposing the  recent results  about   estimation of poses using wheel speeds of \cite{barfoot2014associating,wang2006error} to the context of IMUs. More precisely, our main contributions are as follows:
\begin{itemize}
\item We provide a nontrival extension of the approach and results of \cite{barfoot2014associating} (see also \cite{wang2006error,long2012banana}) which deals with position and orientation (i.e. pose) for wheel sensors based robotics, to   position, orientation plus velocity (i.e. extended pose) when using IMUs, leveraging the  log-linear property of IMU equations of \cite{barrau2017invariant};
\item  We provide an explanation of why the method of IMU on-manifold preintegration  of \cite{forster2017manifold} exists: it is in fact rooted in the group affine property introduced in   \cite{barrau2017invariant}, and the result can be purely expressed by the Lie group $SE_2(3)$. This extends preliminary results of \cite{barrau2018linear} regarding  bias and noise free IMU equations;
\item We provide a more rigorous treatment on the Coriolis effect than \cite{indelman2012factor,indelman2013information},  using the introduced  mathematical framework and a   nontrivial trick, see eq. \eqref{eq:trick}.
\end{itemize}\color{black}
Secondary contributions as follows. First we redemonstrate the   log-linear property of IMU equations  \cite{barrau2017invariant} in discrete time using elementary computations. Then, regarding preintegration we come up with a novel first order development with respect to noise and bias based on Lie exponential coordinates that proves more accurate than the classical Taylor expansion of    \cite{forster2017manifold}. We  derive additional  ``exact''  preintegration formulas when IMU are either noise free, or bias free.

The paper is organized as follows. Section \ref{sec1} is a summary of our preliminary results about noise free and bias free preintegration recently  published in \cite{barrau2018linear}. Section \ref{sec2}  proves the unexpected and novel result that IMU based navigation equations   where Earth rotation is taken into account have the log-linearity property, and hence allow derivation of IMU preintegration formulas in this context. Section \ref{sec3} presents our theory for associating uncertainty with extended poses, with applications to IMU noise propagation. Finally Section \ref{sec4} deals with IMU biases. 
 
\section{A  matrix lie Group  approach to IMU preintegration}\label{sec1}
 We suggest the reader first familiarize with classical Lie groups of robotics, referring to \cite{barfoot2014associating} and ideally to  \cite{barfoot2016state,chirikjian2011stochastic}. 

\subsection{Mathematical Preliminary: the Group $SE_2(3)$}

The special orthogonal group  $SO(3)$ that encodes orientation $\bR$ of a rigid body in space may be modeled as:
$$
SO(3):=\{\bR\in\RR^{3\times 3}\mid \bR^T\bR=I_3,~\det \bR=1\}.
$$
In turn, the set of poses, i.e., position $\bX$ and orientation $\bR$,  may be modeled using the matrix representation of the special Euclidean group
$$
SE(3):=\{\bT=\begin{pmatrix}\bR &\bX\\0_{1,2}&1\end{pmatrix}\in\RR^{4\times 4}\mid (\bR,\bX)\in SO(3)\times\RR^3\}.
$$
Finally to describe \emph{extended poses}, i.e. position $\bX$, velocity $\bV$ and orientation $\bR$, we introduced the following group 
$$SE_2(3):=\{\bT= 
 \begin{pmatrix}\bR&\bV&\bX\\0_{1, 3} &1&0\\0_{1, 3} &0&1\end{pmatrix}\mid (\bR,\bV,\bX)\in SO(3)\times\RR^3 \}, 
$$in \cite{barrau2017invariant}  (see also \cite{barrau2015non}) we called group of ``double direct spatial isometries".  The latter are all matrix Lie groups, embedded in respectively $\RR^{3\times 3}$, $\RR^{4\times 4}$, and $\RR^{5\times 5}$. Matrix multiplication then provides group composition of two elements of $SE_2(3)$. We see   the obtained composition is a natural extension of    poses composition as elementary computations show $(\bR_1,\bV_1,\bX_1)\cdot (\bR_2,\bV_2,\bX_2)=(\bR_1\bR_2,\bR_1\bV_2+\bV_1,\bR_1\bX_2+\bX_1)$.

As in classical Lie group theory, small perturbations of extended poses may be described by elements of the Lie algebra $ \frak{se}_2(3)$. The operator $^\wedge$ turns elements $\boxi:=(\bomega^T,\bv^T,\bx^T)^T\in\RR^9$ into elements of the Lie algebra:
$$ \boxi^\wedge:= \begin{pmatrix}  \bomega \\\bv\\\bx \end{pmatrix}^\wedge=   \begin{pmatrix} (\bomega)_{\times} & \bv & \bx \\ 0_{1,3} & 0 & 0 \\ 0_{1,3} & 0 & 0 \end{pmatrix}$$where   $(\bomega)_{\times}\in\RR^{3\times 3}$ denotes the skew symmetric matrix associated with cross product with $\bomega\in\RR^3$.   The exponential map conveniently maps small perturbations encoded in $\RR^9$ to $SE_2(3)$. For matrix Lie groups it is defined as
$$
\exp(\boxi):=\exp_m(\boxi^\wedge),
$$where 
$\exp_m$ denotes the classical matrix exponential. The following closed form expression may be shown \cite{barrau2017invariant,barrau2015non}:\begin{align}
\exp(\begin{pmatrix}  \bomega \\\bv\\\bx \end{pmatrix})=\begin{pmatrix}  \exp_m((\bomega)_\times)&N(\bomega)\bv&N(\bomega)\bx\\0_{1, 3} &1&0\\0_{1, 3} &0&1\end{pmatrix}\label{exp:map}
\end{align} 
with $N(\bomega)=I_3 +   \frac{1 - \cos(||\bomega||)}{||\bomega||^2}) ((\bomega)_\times)^2 + \frac{ ||\bomega|| -\sin(||\bomega||)}{||\bomega||^3}( (\bomega)_\times)^3$. The Baker-Campbell-Hausdorff (BCH) formula stipulates that for $ \boxi_1,\boxi_2\in\RR^9$ we have $\exp(\boxi_1)\exp(\boxi_2)\approx\exp(\boxi_1+\boxi_2)$ up to second order terms in $ \boxi_1,\boxi_2$. 
Finally, the so-called adjoint operator is defined by analogy to $SE(3)$ as:
\begin{align} 
Ad_{\bT}= \begin{pmatrix}\bR & 0_{3,3} & 0_{3,3} \\ (\bV)_\times \bR & \bR & 0_{3,3}\\ (\bX)_\times \bR & 0_{3,3} & \bR \end{pmatrix}\in\RR^{9\times 9},\label{ad:eq}
\end{align} 
where we conveniently describe it as an operator acting directly on $\RR^9$ instead of  on the Lie algebra $\frak{se}_2(3)$. We have the useful relation that can be considered as a definition:
\begin{align}
\bT \exp(\boxi) \bT^{-1}=\exp(Ad_{\bT}\boxi).
\end{align} 
 
 \subsection{IMU Equations Revisited}

We now  summarize recent results \cite{barrau2018linear}. 
Let $\bR_t$ denote the rotation matrix encoding the orientation of the IMU, and let $\bX_t$ and $\bV_t$ denote  position and velocity of the IMU. Let $\ba_t$ denote the specific acceleration, that is, true acceleration minus gravity vector $\bg$ expressed in the body frame, and $\bomega_t$ the angular velocity expressed in the body frame. The dynamical motion equations on flat Earth write (see \cite{forster2017manifold}):
\begin{align}
\dotex \bR_t=\bR_t(\omega)_\times,\quad\dotex \bV_t=\bg+\bR_t \ba_t,\quad\dotex\bX_t=\bV_t\label{nav:eq}
\end{align}
If we associate a matrix $\bT_t\in SE_2(3)$ to the extended pose $(\bR_t,\bV_t,\bX_t)$, we noticed in \cite{barrau2017invariant}  that \eqref{nav:eq} may be rewritten as
\begin{align}
 \frac{d}{dt} \bT_t= \bW_t \bT_t +  f (\bT_t) +  \bT_t \bU_t,\label{f(ab):eq}
\end{align}
where the various matrices at play write\begin{equation}\label{various:eq1}\begin{aligned}
&\bW_t =
\begin{pmatrix}
0_3 & \bg & 0 \\
 0_{2,3}& 0_{2,1}&0_{2,1}
\end{pmatrix}, \quad
\bU_t=
\begin{pmatrix}
(\bomega_t)_{\times} & \ba_t & 0 \\
 0_{2,3}& 0_{2,1}&0_{2,1}
\end{pmatrix},\\
 & f (\bT_t)=
\begin{pmatrix}
0_3 & 0_{3,1} & \bV_t \\
 0_{2,3}& 0_{2,1}&0_{2,1}
\end{pmatrix}. 
\end{aligned}\end{equation}

It can be easily checked to verify the group affine property:
\begin{dfn}[Group affine dynamics  \cite{barrau2017invariant}]
Let $G$ be a Lie group. Dynamics $\dotex \bT_t=g(\bT_t)$  on $G$ is   \emph{group affine} if it verifies for any couple $\bT,\tilde \bT \in G$ the relation:
\begin{align}
g(\bT\tilde \bT) = g(\bT)\tilde \bT + \bT g(\tilde\bT) -\bT g(Id)\tilde \bT
\label{gpaff}\end{align}
\end{dfn}
This is the key property opening the door to invariant filtering, autonomous error variables, log-linearity and EKF stability  leveraged in e.g. \cite{hartley2019contact,barrau2018invariant}.
The next section summarizes the  links between the latter  formulation of inertial navigation and the theory of preintegration of \cite{lupton2012visual,forster2017manifold}. Notably we have shown in \cite{barrau2018linear} any equation of the form \eqref{f(ab):eq} on a matrix Lie group may in fact be preintegrated.

\subsection{Preintegration of Group Affine Dynamics}

\begin{prop}If $f$   satisfies $f(\bT\tilde \bT) = f(\bT)\tilde \bT + \bT f(\tilde\bT)$, then  dynamics $\frac{d}{dt} \bT_t= g(\bT_t):=\bW_t \bT_t +  f (\bT_t) +  \bT_t \bU_t$ is easily verified to define group affine dynamics. 
\end{prop}
\color{black}

\begin{prop}[\cite{barrau2018linear} Corollary 9]\label{prop::pre-integration}Assuming    $f(\bT\tilde \bT) = f(\bT)\tilde \bT + \bT f(\tilde\bT)$, which is obviously the case with $f$ as in \eqref{various:eq1},  the solution $\bT_t$ at arbitrary  $t$   of  equation \eqref{f(ab):eq} can be written as a function of the initial value $\bT_0$ as:
\begin{align}
\bT_t=\bGamma_t\Phi_t(\bT_0)\bUpsilon_t\label{discrete}
\end{align}where $\bGamma_t,\bUpsilon_t$ are solution to  differential equations involving \emph{only} $\bW_t,\bU_t$, and where $\Phi_t$ only depends on  $t$. Solving the corresponding equations (see \cite{barrau2018linear}) in the particular case of equations    \eqref{nav:eq} on  $SE_2(3)$ with values given by \eqref{various:eq1} yields 
\begin{align}
\Phi_t:
 \begin{pmatrix} \bR&\bV&\bX\\0_{1, 3} &1&0\\0_{1, 3} &0&1\end{pmatrix}\mapsto\begin{pmatrix} \bR&\bV&\bX+t\bV\\0_{1, 3} &1&0\\0_{1, 3} &0&1\end{pmatrix}, \label{phi:eq}\\
\bGamma_t= \begin{pmatrix}I_3&t\bg&\frac{1}{2}\bg t^2\\0_{1, 3} &1&0\\0_{1, 3} &0&1\end{pmatrix}, ~\bUpsilon_t=\begin{pmatrix} \bR_t^\upsilon&\bV_t^\upsilon&\bX_t^\upsilon\\0_{1, 3} &1&0\\0_{1, 3} &0&1\end{pmatrix}, \label{phi2:eq}
\end{align}
where the latter quantities are defined by 
\begin{equation}
\begin{aligned}
\bR_0^\upsilon&=I_3, ~\dotex \bR_t^\upsilon=\bR_t^\upsilon(\omega)_\times,~~ \bV_0^\upsilon&=0,~\dotex \bV_t^\upsilon=\bR_t^\upsilon\ba_t,\\\bV_0^\upsilon&=0,~\dotex \bX_t^\upsilon=\bV_t^\upsilon.
\end{aligned}\label{ODE}\end{equation}
\end{prop}
 
In  \cite{forster2017manifold}, the quantities $\bR_t^\upsilon, \bV_t^\upsilon, \bX_t^\upsilon$ are referred to as the Delta preintegrated measurements and are based solely on the inertial measurements and do \emph{ not} depend on the initial state $\bT_0$. 
This allows one to define  constraints between extended poses at temporally distant key frames based on a \emph{unique} (pre)integration of IMU outputs \eqref{ODE}, no matter how many relinearizations are then used in the optimization scheme.

\section{IMU preintegration with rotating Earth}\label{sec2}

Many applications require accurate localization over long time scales based on accurate inertial sensors. 
To apply factor based optimization techniques to accurate inertial navigation systems requires to take into account  Earth rotation and Coriolis effect. To this date, and to our best knowledge the theory of on-manifold preintegration  cannot handle the  corresponding equations exactly  as the work of \cite{forster2017manifold} is based on  non rotating Earth approximation based equations  \eqref{nav:eq}. 

\subsection{IMU Equations with Rotating Earth are Group Affine}Accounting for Earth rotation,   \eqref{nav:eq}   becomes (see e.g.  \cite{farrell2008aided}):
\begin{equation}
\begin{aligned}
\dotex \bR_t & = -\Omega_{\times} \bR_t + \bR_t(\omega)_\times, \\
\dotex \bV_t & = \bg+\bR_t\ba_t - 2 \Omega_{\times} \bV_t - \Omega_{\times}^2 \bX_t, \quad
\dotex \bX_t & = \bV_t
\end{aligned}\label{nav:eq_Coriolis}
\end{equation}
where $\Omega$ is the Earth rotation vector written in the local (geographic) reference frame. The term $- 2 \Omega_{\times}^2 \bV_t$ is called Coriolis force while the term $- \Omega_{\times}^2 \bX_t$ is called centrifugal force\footnote{To be perfectly accurate, this second term is the varying part of the centrifugal force, which actually writes $-\Omega_{\times}^2 (\bX_t-p_0)$ with $p_0$ a point of the Earth rotation axis. But expanding the parenthesis we obtain a constant term $\Omega_{\times}^2 p_0$ which can be simply added to $g$. And this is already the case: the $g$ we are familiar with (with approximate value $9.81 m.s^{-2}$) is actually the sum of the Newton gravitation force and the centrifugal force due to Earth rotation. Hence the residual term $- \Omega_{\times}^2  \bX_t$.}.   Eq. \eqref{nav:eq_Coriolis} does seemingly not lend itself to application of Prop. \ref{prop::pre-integration}. However, if we  introduce an auxiliary variable:
\begin{align}
\bV_t' = \bV_t + \Omega_{\times} \bX_t\label{eq:trick}
\end{align}
replacing  velocity $\bV_t$,   \eqref{nav:eq_Coriolis}  unexpectedly  simplifies to:
\begin{align*}
\dotex \bR_t & = -\Omega_{\times} \bR_t + \bR_t(\omega)_\times, \\
\dotex \bV_t' & = \bg+\bR_t \ba_t - \Omega_{\times} \bV_t', \quad
\dotex \bX_t   = \bV_t' - \Omega_{\times} \bX_t
\end{align*}
This trick allows embedding the state into a matrix Lie group that fits into the framework of Eq. \eqref{f(ab):eq}:
\begin{align}
 \frac{d}{dt} \bT_t'= \bW_t' \bT_t' +  f (\bT_t') +  \bT_t' \bU_t,
\end{align}
where $f(\cdot)$ and $\bU_t$ are unchanged, and $\bT_t',\bW_t'$ write:
\begin{equation}
 \bT_t'=\begin{pmatrix}
 \bR_t&\bV_t'&\bX_t \\
 0_{1, 3} &1&0\\
 0_{1, 3} &0&1
 \end{pmatrix},
\qquad
\bW_t' =
\begin{pmatrix}
\Omega_{\times} & \bg & 0 \\
 0_{2,3}& 0_{2,1}&0_{2,1}
\end{pmatrix},
\end{equation}
 proving \eqref{nav:eq_Coriolis} are group affine, which is a novel result. 
 
 \subsection{Preintegration with Coriolis Effect}
Using Proposition \ref{prop::pre-integration} we know the equations can be preintegrated. The explicit formulae, given in  \eqref{eq::pre-integration-coriolis} below, may thus be derived along the lines of Section \ref{sec1}.  
\begin{prop}[Preintegration with Coriolis effect]
\label{prop::coriolis}
The IMU equations \eqref{nav:eq_Coriolis} accounting for Coriolis and centrifugal force  with initial state $\bR_0, \bV_0, \bX_0$ write  exactly (no approximation is made):
\begin{equation}
\label{eq::pre-integration-coriolis}
\begin{aligned}
\bR_t & = \bGamma_t^R \bR_0 \bR_t^\upsilon ,\\
\bX_t & = \bGamma_t^x + \bGamma_t^R \bR_0 \bX_t^\upsilon + t \bGamma^R_t \bV_0' + \bGamma^R_t \bX_0, \\
\bV_t & = \bGamma_t^v + \bGamma_t^R \bR_0 \bV_t^\upsilon + \bGamma^R_t \bV_0' - \Omega_{\times} \bX_t,
\end{aligned}
\end{equation}
with $\bV_0'=\bV_0+\Omega_{\times} \bX_0$, and where $\bR_t^\upsilon,\bV_t^\upsilon,\bX_t^\upsilon$ are  the same as in \eqref{ODE} while   $\bGamma^R_t, \bGamma^v_t, \bGamma^x_t$ are defined through the following equations that do \emph{not} involve the state:\begin{equation}
\begin{aligned}
&\bGamma_0^R = I_3, \quad \bGamma_0^v = 0_{3,1}, \quad \bGamma_0^x = 0_{3,1}
;\quad 
\dotex \bGamma_t^R = -\Omega_{\times} \bGamma_t^R , \\&\quad  \dotex \bGamma_0^v = \bg - \Omega_{\times} \bGamma^v_t, \quad \dotex \bGamma_t^x = \bGamma_t^v - \Omega_{\times} \bGamma_t^x.
\end{aligned}\label{ODE21}\end{equation}
\end{prop}

\paragraph{Proof}
After having checked the initial conditions match, we need to check the quantities defined by Eq. \eqref{eq::pre-integration-coriolis} verify Eq. \eqref{nav:eq_Coriolis}. $\dotex \bR_t = -\Omega_{\times} \bR_t + \bR_t \omega_{\times}$ comes easily while we have, using matrix product differentiation rules and the definitions \eqref{ODE} and \eqref{ODE21}, then rearranging terms:
\begin{align*}
\dotex \bX_t & = (\bGamma^v_t + \bGamma_t^R \bR_0 \bV_t^\upsilon + \bGamma^R_t \bV_0') \\
 & - \Omega_{\times} (\bGamma^x_t + \bGamma_t^R \bR_0 \bX_t^\upsilon+ t \bGamma^R_t \bV_0'+ \bGamma^R_t \bX_0)
\end{align*}
which we recognize as $\bV_t$. Now, differentiating $\bV_t$ the same way and using the relation $\dotex \bX_t = \bV_t$ we obtain:
\begin{align*}
\dotex \bV_t = \bg + \bR_t \ba_t -\Omega_{\times} ( \bGamma_t^v +\bGamma_t^R \bR_0 \bV_t^\upsilon + \bGamma^R_t \bV_0' + \bV_t).
\end{align*}
But using the last equality of Eq. \eqref{eq::pre-integration-coriolis} we have $\bGamma_t^v +\bGamma_t^R \bR_0 \bV_t^\upsilon + \Gamma^R_t \bV_0' = \bV_t + \Omega_{\times} \bX_t$ and we end up with:
$$
\dotex \bV_t = \bg + \bR_t \ba_t -2 \Omega_{\times}\bV_t - \Omega_{\times}^2 \bX_t.
$$
$\blacksquare$

The latter novel mathematical   results opens avenues for the   application of factor based optimization methods such as 
GTSAM \cite{dellaertFactor2012} or $g^2o$ \cite{kummerleG2o2011} to real time high performance localization and SLAM based on the use of   high-grade IMUs, along the lines of 
\cite{forster2017manifold}.  

\begin{rem}
\cite{indelman2012factor,indelman2013information}   attacked factor graph based accurate navigation.  The formulas for preintegration with Coriolis effect in the appendix of \cite{indelman2013information},  based on the early approach to preintegration \cite{lupton2012visual} are not presented as exact, and are not indeed, as can be checked for instance propagating Eq. (35) of \cite{indelman2013information}  for one time step, which yields:
$$
v_{j+1}^{L_{j+1}} = R_{L_j}^{L_{j+1}} \left(v_j^{L_j} + R_{b_j}^{L_i} a_j \Delta t + \left[g^{L_i}- 2 \left[\omega_{iL_i}^{L_i} \right]_\times v_i^{L_i}\right] \Delta t \right).
$$
We obtain a term $- 2 \left[\omega_{iL_i}^{L_i} \right]_\times v_i^{L_i}$ in place of the expected $- 2\left[ \omega_{iL_i}^{L_i} \right]_\times v_j^{L_j}$ (index $i$ of $v$ should be $j$): we see Coriolis term is actually approximated by its value at initial time $t_i$.
\end{rem}
 
\section{Associating uncertainty with extended poses}\label{sec3}

 The goal of the present section is twofold. First, it shows how to account for noise in our  approach to preintegration. Then, and more importantly, it provides a generalization of various methods and results of \cite{barfoot2014associating} devoted to $SE(3)$ to the case of extended poses of  $SE_2(3)$. This extension is independent from the theory of preintegration and is a contribution in itself.  It is not trivial as  even using the recently introduced $SE_2(3)$ group,  IMU propagation  is not amenable to Lie group   compounding $\bT_{k+1}\exp(\boxi_{k+1})=\bT_{k} \exp(\boxi_{k})\bar \bT\exp(\boxi)$ as considered in  \cite{barfoot2014associating}. The unexpected log-linearity property of \cite{barrau2017invariant}, that we rederive more simply below,   plays a key role.

\subsection{Associating uncertainty with elements of $SE_2(3)$}

Using the exponential map of $SE(3)$ to describe statistical dispersion of poses  has been often advocated.  In the robotics community, early attempts date back to \cite{smith1986representation}, and references \cite{chirikjian2011stochastic,barfoot2016state,chirikjian2014gaussian,barfoot2014associating,BayesianLieGroups2011,diemer2015invariant,barrau2013intrinsicp,hertzberg2013integrating,park2008kinematic,long2012banana,wang2006error} revolve around those ideas. Gaussians in exponential coordinates are also referred to as  concentrated Gaussians \cite{bourmaud2013discrete}.  
We define a (concentrated) Gaussian  on $SE_2(3)$ as 
\begin{align}
\bT =\bar\bT \exp(\boxi)\label{error:rep1},
\end{align}
where $\bar \bT$ is a noise free ``mean" of the distribution and $\xi\sim\mathcal N(0,\bSigma)$ is a  multivariate Gaussian in $\RR^9$. Each  component of $\boxi=(\bomega^T,\bv^T,\bx^T)^T$ corresponds to a degree of freedom.


\subsection{Propagation of Errors through Noise Free IMU Model}

We now come back to the widespread flat Earth model \eqref{nav:eq} and consider discrete time, leveraging  formula  \eqref{discrete}. In discrete time with time step $\Delta t$, denoting $\bGamma_k:=\bGamma_{\Delta t}$, $\Phi:=\Phi_{\Delta t}$ and $\bUpsilon_k:=\bUpsilon_{\Delta t}$, we  use  \eqref{discrete} to get indeed \begin{align}\bT_{k+1}=\bGamma_k\Phi(\bT_k)\bUpsilon_k\label{discrete2}.\end{align} 

\begin{rem}
Contrary to \cite{forster2017manifold} our matrix group based preintegration formula \eqref{discrete} is an exact discretization of  \eqref{nav:eq}.  However it involves \eqref{ODE} that needs to be numerically solved (but the beauty of preintegration is that it needs be solved only once, and then  the same formula  \eqref{discrete} may be used over and over for various initial conditions). \color{black}
As IMU measurements come in discrete time, albeit at a high rate, we may call $\Delta t$ the discretization step and assume   $\ba_t$, $\bomega_t$ to be constant over time  intervals of small size $\Delta t$, along the lines of \cite{forster2017manifold} (in IMUs this assumption is inevitable, and its negative impact mitigated by the high frequency of measurements).\color{black} The solution  to  \eqref{discrete} is based on   \eqref{ODE} whose solution then writes $ \bR_{t+\Delta t}^\upsilon= \bR_{t}^\upsilon\exp_m((\omega_t)_\times \Delta t)$,  $\bV_{t+\Delta t}^\upsilon=\bV_{t}^\upsilon+ \bR_{t}^\upsilon\ba_t\Delta t$, and $\bX_{t+\Delta t}^\upsilon=\bX_{t}^\upsilon+ \bV_{t}^\upsilon\Delta t+ \bR_{t}^\upsilon\ba_t\Delta t^2$ under the  approximation of $ \bR_{t}^\upsilon$ being also constant over $\Delta t$ in the second equation. 
\end{rem}
We are now in a position to derive a preliminary yet    remarkable  result regarding   propagation through noise free IMU equations of a concentrated Gaussian \eqref{error:rep1}. 
\begin{prop}[Extended pose error propagation]Let $\bT_k =\bar\bT_k \exp(\boxi_k)$ where both $\bT_k $ and $\bar\bT_k$ evolve through noise free   model  \eqref{discrete2}. The propagation of discrepancy $\boxi_k\in\RR^9$ between $\bar\bT_k$ and $\bT_k$ writes $\boxi_{k+1}=Ad_{\bUpsilon^{-1} } F\boxi_k$ with  $F(\bomega,\bv,\bx)^T:=(\bomega,\bv,\bx+\Delta t \bv)^T$, i.e.,   \begin{align} 
\bT_{k+1}:=\bGamma_k  \Phi (\bar \bT_k \exp(\boxi_k))\bUpsilon_k =\bar \bT_{k+1}\exp(Ad_{\bUpsilon_k^{-1} } F\boxi_k) 
\label{magic:formula}\end{align}\label{bananaprop}
\end{prop}
 \paragraph{Proof}We have
$\bT_{k+1}=\bGamma  \Phi (\bar \bT_k \exp(\boxi_k))\bUpsilon \\
=\bGamma  \Phi (\bar \bT_k)\Phi ( \exp(\boxi_k))\bUpsilon $ where we used   $\Phi ( \bT_1 \bT_2)=\Phi ( \bT_1  )\Phi (   \bT_2)$ as easily shown  from \eqref{phi:eq}. Thus
 $
\bT_{k+1} =\bGamma  \Phi (\bar \bT_k)\bUpsilon\bUpsilon^{-1} \Phi ( \exp(\boxi_k))\bUpsilon 
 =\bar \bT_{k+1}Ad_{\bUpsilon^{-1} }(\Phi ( \exp(\boxi_k))).
$ 
Using the expression  \eqref{exp:map}  and \eqref{phi:eq} we see that
\begin{align}
\Phi(\exp(\boxi))=\exp(F\boxi)\label{line16}
\end{align}
where $F(\bomega,\bv,\bx)^T=(\bomega,\bv,\bx+\Delta t \bv)^T$. Using that matrix exponential commutes with conjugation we get $Ad_{\bUpsilon^{-1} }  ( \exp(F\boxi))=\exp(Ad_{\bUpsilon^{-1} } F\boxi)$, proving the result. $\blacksquare$
 
We have just proved in discrete time using elementary means the log-linear property of \cite{barrau2017invariant} dealing  with continuous time. It proves the interest of  error representation \eqref{error:rep1} using exponential coordinates in $SE_2(3)$, as \eqref{magic:formula} shows the errors $\boxi_k$ expressed using those coordinates in  
$\RR^9$ propagate \emph{linearly} in exponential coordinates through IMU equations \eqref{discrete} and hence \eqref{nav:eq},  as \eqref{ad:eq} ensures that $Ad_{\bUpsilon^{-1} } F\in\RR^{9\times 9}$.   

\begin{rem}
\cite{barfoot2014associating} considers    propagation by   compounding and proves noise free propagation    $  \bT_{k+1} =  \bGamma_k\bT_{k} \bUpsilon_k$, in place  of the more sophisticated  dynamics \eqref{discrete2},   preserves statistical distributions \eqref{error:rep1},  see Eq.  (26) in Reference \cite{barfoot2014associating}. It also  explains why   dispersion of wheeled robots is  banana shaped   \cite{long2012banana,chirikjian2011stochastic,barfoot2016state,chirikjian2014gaussian}. We recover the result in a more complex case, as $\Phi$ doesn't boil down to   simple multiplications. What  saves the day, though,   is that $\Phi$  is a group automorphism, i.e.   $\Phi ( \bT_1 \bT_2)=\Phi ( \bT_1  )\Phi (   \bT_2)$, see also \cite{barrau2018linear}.\color{black}\label{rem1}
\end{rem}

\subsection{IMU Noise Model}\label{justify:sec}

In practice IMU actually measure   $\tilde \ba_k:=\ba_k-\bb_k^a-\bbeta_k^a$,  and  $\tilde \bomega_k:=\bomega_k-\bb_k^g-\bbeta_k^g$, with $\bb_k=(\bb_k^g,\bb_k^a)\in\RR^6$ the gyrometers and accelerometers biases,  and where $\bbeta$ represents sensor noises.  For now, let's  assume  $\bb=0$, as biases will be the focus of Section \ref{sec4}. Integrating   \eqref{ODE} for one time step yields to the first order  in $\Delta t$:
\begin{align}
\bUpsilon=\begin{pmatrix} \exp(\tilde\bomega\Delta t)_\times &\tilde\ba\Delta t&0_{3, 1}\\0_{1, 3} &1&0\\0_{1, 3} &0&1\end{pmatrix}.\label{backtofuture}
 \end{align}A simple matrix multiplication proves that this implies the following first order relation between factor $\bUpsilon$ associated to noisy  inertial increments and $\bar\bUpsilon$ associated to noise free ones\begin{align}
\bUpsilon\approx\bar\bUpsilon\exp(\bbeta^g\Delta t,\bbeta^a\Delta t,0):=\bar\bUpsilon\exp(\bbeta_k),
 \end{align}
and we let $\bbeta_k:=(\bbeta_k^g\Delta t,\bbeta_k^a\Delta t,0)\in\RR^9$. Recalling   \eqref{discrete2}, we then have in the presence of IMU noise the motion model:
\begin{align}\bT_{k+1}=\bGamma_k\Phi(\bT_k) \bar\bUpsilon_k\exp(\bbeta_k),\label{discrete3}\end{align}
and we now seek to describe how the distribution \eqref{error:rep1}, i.e., its associated parameters $\bar\bT,\bSigma$, propagate through \eqref{discrete3}.

\subsection{Propagation   with Noisy IMU: an Exact Formula}

The theory allows deriving a novel result describing error accumulation over time.
As show by \eqref{discrete3} the noisy version of  \eqref{discrete2} has the following form:
\begin{align}
\bT_{k+1} = \bGamma_k \Phi(\bT_k) \bUpsilon_k \exp(\bbeta_k),\label{noisy:eqq}
\end{align}
with $\bbeta_i's$ independent centered Gaussian noises.  The group affine property of the $SE_{2}(3)$ embedding allows deriving a    novel result describing error accumulation over time.
\begin{prop}[Extended pose error accumulation]
\label{prop::acc}Referring to \eqref{error:rep1} let us write $\bT_k$ as $\bT_k=\bar\bT_k\exp(\boxi_k)$ where $\bar\bT_k$ is propagated through noise free equations \eqref{discrete2} i.e., $\bar\bT_{k+1}=\bGamma_k \Phi(\bar\bT_k ) \bUpsilon_k$. 
Let $\bF_k:=Ad_{\bUpsilon_k^{-1} } F\in\RR^{9\times 9}$ and  $\bF_i^k:=\Pi_{j=i}^{k-1} \bF_j$. We have the recursive formula\begin{equation}
\label{eq::one-step}
\exp(\boxi_{k+1}) = \exp(\bF_k \boxi_k) \exp(\bbeta_k)
\end{equation}leading to the following exact formula:
\begin{equation}
\label{eq::prod}\boxed{
\exp(\boxi_{k}) = \exp(\bF_0^{k-1} \boxi_0) \cdot \prod_{i=0}^{k-1} \exp(\bF_{i+1}^{k-1} \bbeta_i) . }
\end{equation}
\end{prop}

 \paragraph{Proof}
Juste before \eqref{line16} we proved
$
\exp(\boxi_{k+1})   
 = Ad_{\bUpsilon_k^{-1} }(\Phi ( \exp(\boxi_k))) 
$ for noise free model \eqref{discrete2}. With noisy model \eqref{noisy:eqq} we have along the same lines
\begin{align}\label{line122}
\exp(\boxi_{k+1}) = Ad_{\bUpsilon_k^{-1} }(\Phi ( \exp(\boxi_k)))\exp(\bbeta_k).
\end{align}
Let $\Psi_k$ denote $Ad_{\bUpsilon_k^{-1} }\circ\Phi$. At \eqref{line16} and just after we proved \begin{align}\label{line123}\Psi_k(\exp(\boxi)):=Ad_{\bUpsilon_k^{-1} }\circ \Phi(\exp(\boxi))= \exp(Ad_{\bUpsilon_k^{-1} }F\boxi)\end{align}hence $\Psi_k(\exp(\boxi))=\exp(\bF_k\boxi)$ 
readily proving \eqref{eq::one-step}. Moreover we proved $\Phi(\bT_1\bT_2)=\Phi(\bT_1)  \Phi(\bT_2)$, i.e. $\Phi$ is an automorphism. The adjoint is well-known to be an automorphism also. As the composition of automorphisms is an automorphism,   $\Psi_k$ satisfies the same property. We have
\begin{align*}
\exp(\boxi_{2}) & = Ad_{\bUpsilon_1^{-1}} \circ \Phi\left(Ad_{\bUpsilon_0^{-1}} \circ \Phi(\exp(\boxi_0)\exp(\bbeta_0) \right) \exp(\bbeta_1) \\
 & = \Psi_1\bigg( \Psi_0(\exp(\boxi_0))\exp(\bbeta_0) \bigg) \exp(\bbeta_1)
 \\
 & = \Psi_1(\Psi_0(\exp(\boxi_0))\Psi_1(\exp(\bbeta_0) )) \exp(\bbeta_1)
  \\
 & =  \exp(\bF_1\bF_0\boxi_0)) \exp(\bF_1\bbeta_0) ) \exp(\bbeta_1)
\end{align*}and \eqref{eq::prod} is proved by recursion along the same lines. $\blacksquare$
 
This result is remarkable: having a simple closed-formula for the error propagation is not usual in nonlinear state estimation. A first   application is  that    if the initial error $\boxi_0$ and the noises $\bbeta_i$ are centered on zero then the propagated error is centered up to the third order w.r.t. the standard deviation of $\boxi_0$ and the $\bbeta_i$'s.  This may be  proved along the lines of \cite{barfoot2014associating}  that deals with  the simpler case of compounding.

Albeit unusual to be able to come up with exact formulas such as \eqref{eq::prod}, the formula  is yet nonlinear. The evolution of $\boxi_k$'s covariance may be approximated as follows. 

\begin{prop}[IMU noise propagation]Consider a sequence of uncertain extended poses, modeled as $\bT_k=\bar\bT_k\exp(\boxi_k)$ with $\boxi_k\sim\mathcal N(0,\bSigma_k)$. Using the BCH formula to the first order in \eqref{eq::one-step} readily provides an approximation of uncertainty propagation through noisy IMU model \eqref{discrete3} as \begin{align}\boxed{ \bar\bT_{k+1}=\bGamma_k \Phi(\bar\bT_k ) \bUpsilon_k,\quad 
\bSigma_{k+1}=\bF_k\bSigma_k\bF_k^T+\bSigma_{\bbeta}.}\label{riccati}
\end{align}
\end{prop}
Up to the first order, the obtained Riccati equation agrees with the results of  \cite{forster2017manifold}, see   appendix therein. However,   higher-order formulas are different and shall be explored in future work in a similar way to \cite{barfoot2014associating,chirikjian2011stochastic}, but on $SE_2(3)$. 

Riccati equation \eqref{riccati}   only provides an approximation to the true formula \eqref{eq::prod}. However,  Proposition \ref{bananaprop}  shows propagation of concentrated Gaussians is exact when sensors are unnoisy, which is a good indication of accuracy in the case where noise  is present but  moderate.  This is also supported by the simple numerical experiment of Figure \ref{MC:fig} where we see true and computed dispersions  match indeed. 
{\center{
\begin{figure}
\includegraphics[width=\textwidth]{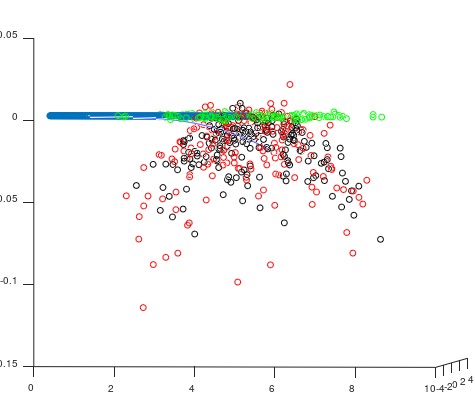}
\caption{Numerical experiment to support uncertainty  propagation model \eqref{riccati} coupled with uncertainty representation \eqref{error:rep1}. The initial extended pose is null and known, and the IMU  moves nominally to the right at constant translational speed (blue line).   Noisy IMU measurements generate a dispersion of the belief at the endpoint. We generate point clouds at the trajectory endpoint based on Monte-Carlo simulations. Black dots represent true dispersion under noisy equations \eqref{nav:eq}. Red dots are generated through our exponential uncertainty model \eqref{error:rep1} for extended pose propagation with parameters computed via \eqref{riccati}. Finally green dots are generated using the  endpoint distribution computed by a standard (multiplicative)  EKF: we see linearization implies the  assumed dispersion lies within a horizontal plane. However, the true distribution (black) is ``banana'' shaped in 3D, as already observed mainly in the case of poses in 2D for wheeled robots \cite{long2012banana,chirikjian2011stochastic,barfoot2016state,chirikjian2014gaussian,barfoot2014associating}, and \eqref{riccati} captures this effect and agrees with ground truth. }\label{MC:fig}
\end{figure} 
}}

\section{Impact of biases in exponential coordinates for preintegration}\label{sec4}

In this section, we compute first-order bias correction using our representation of errors   based on exponential coordinates on $SE_2(3)$. First,  our matrix formalism allows for more elementary computations than the first order expansions that can be found in the Appendix of  \cite{forster2017manifold}. Second   our theory yields slightly more accurate Jacobians for first-order bias correction in the theory of preintegration.  

\subsection{Theory}

Consider full IMU measurements $\tilde \ba_k:=\ba_k-\bb_k^a-\bbeta_k^a$ and $\tilde \bomega_k:=\bomega_k-\bb_k^g-\bbeta_k^g$, and let us  ignore the noise and focus only on the biases. In the context of smoothing,  given a bias update $\bar \bb\leftarrow \bar \bb+\db$, one needs to compute how the preintegrated quantities change. Assume we have computed the extended pose $\bT_k(\bar \bb)$ at time $k$ corresponding to bias $ \bar \bb$ and let $\bT_k(\hat \bb)$ denote the  extended pose associated to new bias estimation $\hat \bb:=\bar \bb+\db$ with   $\db=(\db^g,\db^a)\in\RR^6$. Building upon our exponential representation of errors on $SE_2(3)$  \eqref{error:rep1}  in a stochastic context, we define the discrepancy between the associated extended poses as $\bd_k\in\RR^9$, i.e., 
\begin{align}\bT_k(\hat \bb)=\bT_k(\bar \bb)\exp(\bd_k),\label{bias:discrep}
 \end{align}
 and we seek how this  correction evolves with time.  
Denote $\bUpsilon(\bb)$ the quantity obtained by replacing $(\tilde\bomega,\tilde\ba)$ with $(\tilde\bomega+\bb^g,\tilde\ba+\bb^a)$ in   \eqref{backtofuture}. Neglecting $O(\Delta t^2)$ terms yields
 \begin{align}
\bUpsilon(\hat \bb)\approx\bUpsilon(\bar \bb)\exp(\db^g\Delta t,\db^a\Delta t,0).
\label{approx2:eq}
 \end{align}Thus using \eqref{discrete2},  \eqref{bias:discrep} and \eqref{approx2:eq} we   get
 \begin{align}
\bT_{k+1}(\hat \bb) &=\bGamma_k\Phi  ( \bT_{k}(\hat \bb))\bUpsilon_k(\hat \bb) \\&=\bGamma_k  \Phi (  \bT_k(\bar \bb) \exp(\bd_k))\bUpsilon_k(\bar \bb) \exp(\bar{\db} ).
 \end{align}
where   $\bar\db:=(\db^g\Delta t,\db^a\Delta t,0)\in\RR^9$. Using \eqref{magic:formula} we get
 \begin{align}
\bT_{k+1}(\hat \bb )&=  \bT_{k+1}(\bar \bb)\exp(Ad_{\bUpsilon_k(\bar \bb)^{-1} } F\bd_k)\exp(\bar{\db})\\
&\approx  \bT_{k+1}(\bar \bb)\exp(Ad_{\bUpsilon_k (\bar \bb)^{-1}} F\bd_k+\bar{\db}) ,\label{approx:eq}
 \end{align}
where we used the BCH formula. We have thus proved the discrepancy $\bd_k$ in the sense of \eqref{bias:discrep} between extended poses respectively associated to biases $\bar\bb$ and $\hat\bb=\bar\bb+\db$ satisfies 
 \begin{align}\boxed{\bd_{k+1}=Ad_{\bUpsilon_k (\bar \bb)^{-1}} F\bd_k+\bar{\db}}\label{biascor}
 \end{align}
We see the only approximation\footnote{besides the Euler  approximation \eqref{approx2:eq} justified by small $\Delta t$.} comes in at line \eqref{approx:eq}. Note \eqref{biascor} may rewrite  
$
\bd_k=J_k\bar{\db},~\text{where}~J_{k+1}=Ad_{\bUpsilon_k(\bar \bb)^{-1} } FJ_k+I_{9,6}$ and where $I_{9,6}$ is a $6\times 6$ identity matrix concatenated with a $3 \times 6$ matrix of zeros, and we let  $J_0$ be the identity.    
\begin{rem}
Neglecting terms in $\Delta t^2$ in \eqref{approx2:eq} alleviates computations but is not fully accurate. For correct expansion w.r.t.  $ \bar{\db}$ the diagonal identity blocks $1:3 \times 1:3$ and $4:6 \times 4:6$ of matrix $I_{9,6}$ should be replaced with $\Delta t D$ and $\Delta t \exp_m((\bomega_t)_{\times} \Delta t)$ respectively, where $D$ is defined by the expansion $\exp_m((\bomega+u)_{\times})=\exp_m(\bomega)[I_3+(Du)_{\times}+\circ(u)]$.
\end{rem}The following result is already true for classical Taylor expansion of \cite{forster2017manifold}, although it seems to have been  noticed only in  \cite{martinelli2014closed}.
\begin{prop}
In absence of gyro bias both formulas \eqref{approx2:eq} and \eqref{approx:eq} are exact, meaning we have exact pre-integration where    accelerometer bias is fully modeled. 
\end{prop}

%
%
%

\subsection{Numerical comparison}

We showed the exponential mapping on $SE_2(3)$ more closely reflects   uncertainty when using noisy IMUs. In turn, one may wonder if using the exponential to model bias correction \eqref{bias:discrep} improves accuracy. The answer turns out to be positive, although the improvement is rather slight.

\begin{figure}
\includegraphics[width=.8\columnwidth]{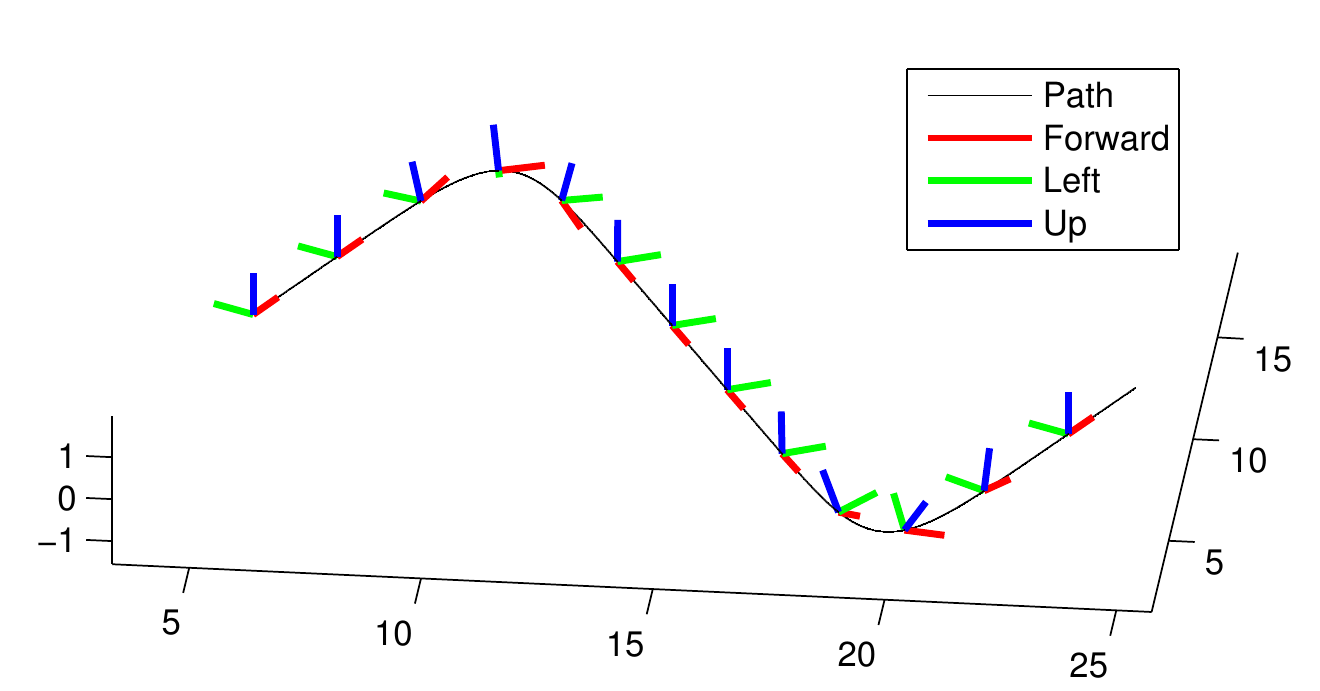}
\caption{Simulated trajectory}
\label{fig::traj}
\end{figure}

We set a simple simulation where a UAV follows a 3D trajectory  (see Fig. \ref{fig::traj}) while recording IMU measurements, and storing preintegrated factors, each covering a duration $T$. The original sampling frequency is 100Hz, so   $T=0.01$ means all measurements are stored. Then we sample values of the gyro and accelerometer bias and compute the difference between preintegrated factors obtained re-integrating the IMU increments and pre-integrated factors obtained through the two first-order expansion,   as in standard preintegration theory \cite{forster2017manifold} and exponential as in \eqref{bias:discrep}, \eqref{biascor}. Results for the velocity and position components of the pre-integrated factors are displayed on Table \ref{tab::res} for a bias   corresponding to a low-cost IMU. We see the exponential mapping of our group theoretic approach tends to improve velocity accuracy of the first-order expansion. As  the errors between the preintegrated factors and their first-order approximation are very small for a standard pre-integration time of $T=1s$, we conclude that regarding bias correction exponential mapping may prove useful   in specific situations such as long term preintegration.

%

\begin{table}
\centering
\caption{Higher-order errors for a low-cost IMU ($1^\circ/s - 100 mg$)}
\label{tab::res}
\begin{tabular}{|c|c|c|c|c|c|c|}
\hline
IMU & \multicolumn{3}{c|}{Velocity RMS (m/s)} & \multicolumn{3}{c|}{Position RMS (m)} \\
\hline
$T$ & $1s$ & $10s$ & $60s$& $1s$ & $10s$ & $60s$ \\
\hline
 Classical & $6,9.10^{-3}$ & $1,25$ & $151,5$ & $0,0016$ & $3,66$ & $2405,8$ \\
\hline
 Proposed & $9,375.10^{-4}$ & $0,37$ & $70,5$ & $0,0015$ & $2,57$ & $2126,5$ \\
\hline
\end{tabular}
\end{table}

\section{Conclusion}
We showed the properties of $SE_2(3)$ allow transposing the rather recent results  about   estimation of poses using wheel speeds  to the context of IMUs. 
Moreover, the framework provides an elegant mathematical approach that brings further maturity to the theory of preintegration on manifolds. It unifies flat and rotating Earth IMU equations within a single framework, hence providing extensions of the theory of preintegration to  the context of high grade IMUs and opening up for novel implementations of factor graph based methods to high precision (visual) inertial navigation systems.


\end{document}